\documentclass[pdflatex,sn-mathphys-num]{sn-jnl}

\usepackage{graphicx}%
\usepackage{multirow}%
\usepackage{amsmath,amssymb,amsfonts}%
\usepackage{amsthm}%
\usepackage{mathrsfs}%
\usepackage[title]{appendix}%
\usepackage{xcolor}%
\usepackage{textcomp}%
\usepackage{manyfoot}%
\usepackage{booktabs}%
\usepackage{algorithm}%
\usepackage{algorithmicx}%
\usepackage{algpseudocode}%
\usepackage{listings}%

\usepackage{graphicx}
\usepackage{array}

\theoremstyle{thmstyleone}%
\theoremstyle{thmstyletwo}%

\theoremstyle{thmstylethree}%

\begin{document}

\title{A Comprehensive Review on Understanding the Decentralized and Collaborative Approach in Machine Learning}


\author*[1]{\fnm{Sarwar} \sur{Saif}}\email{saifmu6@gmail.com}

\author[2]{\fnm{Md Jahirul} \sur{Islam}}\email{jimonir004@gmail.com}

\author[3]{\fnm{Md. Zihad Bin} \sur{Jahangir}}\email{zihad.bscincse@gmail.com}

\author[4]{\fnm{Parag} \sur{Biswas}}\email{text2parag@gmail.com}

\author[5]{\fnm{Abdur} 
\sur{Rashid}}\email{rabdurrashid091@gmail.com}

\author*[6]{\fnm{MD Abdullah Al} \sur{Nasim}}\email{nasim.abdullah@ieee.org}

\author[7]{\fnm{Kishor} \sur{Datta Gupta}}\email{kgupta@cau.edu}

\affil[1]{\orgdiv{Department of Business Sciences, Humanities and Social Sciences}, \orgname{University of Tsukuba}, {\city{Tsukuba}, \country{Japan}}}

\affil[2, 3, 6]{\orgdiv{Research and Development Department}, \orgname{Pioneer Alpha},  \orgaddress{{\city{Dhaka}, \country{Bangladesh}}}}

\affil[4, 5]{\orgdiv{MSEM Department}, \orgname{Westcliff university},  \orgaddress{{\city{California}, \country{United States}}}}

\affil[7]{\orgdiv{Department of Computer and Information Science}, \orgname{Clark Atlanta University}, {\city{Georgia}, \country{USA}}}


\abstract{ The arrival of Machine Learning (ML) completely changed how we can unlock valuable information from data. Traditional methods, where everything was stored in one place, had big problems with keeping information private, handling large amounts of data, and avoiding unfair advantages. Machine Learning has become a powerful tool that uses Artificial Intelligence (AI) to overcome these challenges. We started by learning the basics of Machine Learning, including the different types like supervised, unsupervised, and reinforcement learning. We also explored the important steps involved, such as preparing the data, choosing the right model, training it, and then checking its performance. Next, we examined some key challenges in Machine Learning, such as models learning too much from specific examples (overfitting), not learning enough (underfitting), and reflecting biases in the data used. Moving beyond centralized systems, we looked at decentralized Machine Learning and its benefits, like keeping data private, getting answers faster, and using a wider variety of data sources. We then focused on a specific type called federated learning, where models are trained without directly sharing sensitive information. Real-world examples from healthcare and finance were used to show how collaborative Machine Learning can solve important problems while still protecting information security. Finally, we discussed challenges like communication efficiency, dealing with different types of data, and security. We also explored using a Zero Trust framework, which provides an extra layer of protection for collaborative Machine Learning systems. This approach is paving the way for a bright future for this groundbreaking technology.}

\keywords{Machine Learning, Decentralized Learning, Federated Learning, Zero Trust Framework}



\maketitle

\section{Introduction}\label{sec1}

Machine Learning is a sub-discipline of artificial intelligence that has fundamentally changed processes of interacting with technology, allowing us to extract vital information from large quantities of data. ML algorithms can train computers to identify complex patterns human observers cannot discern. This ability enables the creation of machines that anticipate situations, make judgment calls like a person, and learn from their mistakes or successes. Conventionally, ML was organized using a centralized paradigm, where enormous amounts of data had to be aggregated in one place to feed into training \cite{JordanandMitchell2015}.
Centralized training provokes a security risk thanks to the potential for secret information leaks, primarily for computational reasons, as choosing a single computer or group of computers to train a model generates the hazards of biases and discrimination \cite{McMahanetal2017}. This has been nascent in machine learning paradigms and needs to be more frequently considered relevant. Thus, there is a uniform tendency toward decentralization and cooperative strategies. Decentralized Machine Learning is a development that allows the processing of learning algorithms to be scattered through networks of devices or nodes. This necessity does not just offer privacy for training with the given data; it further creates scalability \cite{Yangetal2019}.
Federated Learning, a specific model of DML, allows training a model on separate customer devices devoid of each other’s data being in a central location, stopping data from leaving the devices and obscuring obscure data \cite{Lietal2020}. Alternatively, FL and additional decentralized training paradigms can be utilized to learn over the network in numerous fields. Perhaps what is needed is conserving the secrecy of information in the medico industry, or it is concealing information on the phone from the holders and interested factions while aiming to personalize the performance of various device attributes \cite{Kairouzetal2021}.
However, although decentralization answers critical business issues, it also engenders problems. These include heavier communication and security concerns involving all the machines and devices within a network. Hence, the following chapter begins by presenting what machine learning is and then examines the restrictions of the centralized strategy. We will elaborate on decentralized and unified machine learning approaches, illustrating their purpose and explaining implementation measures by revealing the intrinsic and external motivations. The second part will then delve into complexity, offer possible solutions for machine-decentralized learning, and then explain the singularity scope via the dominated by Zero Trust security \cite{Stafford2020}.

The contributions of this research paper are summarized below:

\begin{enumerate}
    \item Discussing the privacy concerns in terms of applying ML in numerous sectors. Understanding the processing of data without sharing sensitive information. Stating the scaling and efficiency is also another important part of this paper. 

    \item Demonstrating the applications of federated learning for healthcare and finance systems to solve critical problems. Bias and Fairness in distributed learning is also another concern of this manuscript.

    \item Moreover, this paper suggests incorporating frameworks that work collaboratively with other frameworks that follow the zero-trust mechanism.

\end{enumerate}

\section{Literature Review}
There is a breach of privacy when centralized machine learning systems aggregate vast amounts of data, which is an issue on its own. Distributed and decentralized networks solve this problem as no such aggregation entity \cite{Abadietal2016, McMahanetal2017}. The problem is that there has been a heightened concern for citizens in recent years, which goes further with more stringent privacy legislation. On the other hand, central machine-driven models can no longer deal with data expansion because they are severely lagging on the expansion of datasets, and it has resulted in cases of limited data acquisition, processing bottlenecks, and financial constraints \cite{ChenLin2014, Lietal2020}—their productivity with regards to the significant data age not meeting the requirements. Also, predispositions typically embedded in the data that centralized models can amplify must be identified and clarified. Like these other methods, failure to do this can result in unequal or unjust outcomes; therefore, ethical and responsible AI development is essential.
There is a rising movement towards decentralized and collaborative machine learning systems as a reaction to these restrictions. Potential answers are provided by decentralized machine learning (DML), which distributes processing across networks, allows data to stay locally, and lessens the need for centralized aggregation \cite{Konecnyetal2016, Lietal2020}. In addition to improving privacy, this method can help with scalability problems. Federated Learning (FL) is an important cooperative method that permits training models on distributed data without compromising privacy; participants maintain data ownership while sharing knowledge and potentially mitigating biases through more diverse data \cite{Konecnyetal2016, McMahanetal2017}. However, decentralization and collaboration introduce their own challenges, such as communication overhead impacting efficiency \cite{Lietal2020}.  Ensuring robust security in decentralized systems and guaranteeing consistent performance across heterogeneous data and devices remain active areas of research \cite{Hitajetal2017, Lietal2020, Kairouzetal2019}.
The method of zero-trust security, with no transaction of implicit trust and assumption of continued verification, penetrates machine learning systems where Rose et al. (2020) have recommended it. Typical ML strategies often accept a certain degree of trust towards data sources and participants, leading to their vulnerability to threats like model poisoning (attacks on learning models that alter the standard trustworthiness and lead to breaches in data privacy). Adopting the zero-trust model’s principles and truncating the risks for ML come into the picture \cite{Hitajetal2017, Lietal2021}. Focused investigations are being conducted on several methods, including differential privacy, which adds controlled noise to individual data points and maintains statistical properties without revealing their accurate inputs \cite{DworkRoth2014}, secure multi-party computation, which enables collaboration in such scenarios without the participants revealing their private inputs \cite{Lindell2020}, and homomorphic encryption, that enables the performance of computations directly on encrypted data. Adding this concept increases the performance of federated learning systems and decentralized ML. However, it is essential to note that the process may lead to slowdowns in computation or may impact the performance of the models; therefore, the need for more research in optimizing their use in the ML context keeps emerging.

Authors have summarized the potential challenges from the literature in Table \ref{tab:lit_review_summary}. From there the potential challenges have been understood and the pros along with the cons are stated.

\begin{table}[h!]
\centering
\caption{Literature Summary from State-of-the-art Papers}
\begin{tabular}{|p{3.5cm}|p{5cm}|p{3cm}|}
\hline
\textbf{Topic} & \textbf{Solutions in Decentralized ML} & \textbf{Potential Challenges} \\
\hline
\textbf{Privacy} & Local memory and data preservation & Zero-trust security to prevent model poisoning and privacy threats \\
\hline
\textbf{Scalability} & Distributed processing improves scalability & Communication overhead and consistency across heterogeneous data \\
\hline
\textbf{Bias \& Ethical Concerns} & FL allows model training with diverse, unbiased data & Maintaining ethical standards in decentralized ML \\
\hline
\textbf{Techniques for Security} & Differential Privacy, Secure Multi-Party Computation, Homomorphic Encryption & Computational slowdowns and performance impact \\
\hline
\textbf{Emerging Approaches} & Zero-trust principles to enhance decentralized ML security & Ongoing research required to optimize security and performance trade-offs \\
\hline
\end{tabular}
\label{tab:lit_review_summary}
\end{table}

\section{Methodology}
The research methodology of the “Introduction to Machine Learning and Decentralized and Collaborative Machine Learning" follows a systematic approach to identify, select, and analyze relevant literature. This qualitative review focuses on decentralized machine learning (DML) and federated learning (FL), emphasizing privacy, scalability, and security.
The literature search was conducted using multiple academic databases and reputable sources to ensure a comprehensive review of the current state of decentralized and collaborative machine learning. The primary databases included IEEE Xplore, ACM Digital Library, Google Scholar, and PubMed. Additionally, conference proceedings from major ML and AI conferences such as NeurIPS, ICML, and AISTATS were reviewed, along with relevant white papers and technical reports from reputable institutions and companies.
We then carefully selected search terms and keywords to capture the core concepts and latest developments in decentralized and federated learning. These included terms like "Decentralized machine learning," "Federated learning," "Collaborative machine learning," "Privacy-preserving machine learning," "Distributed learning algorithms," and "Zero Trust security model in machine learning." Our literature search focused on publications from the last ten years to ensure the inclusion of recent and relevant studies, while also considering earlier seminal works for their foundational insights.
Furthermore, we established specific inclusion criteria for the review, encompassing peer-reviewed journal articles, conference papers, and reputable white papers that addressed decentralized and federated learning methodologies, applications, and security concerns. We also prioritized studies that provided empirical data or substantial theoretical contributions. Exclusion criteria included non-English articles, publications focusing solely on centralized machine learning, and studies lacking methodological rigor or detailed experimental results.
The selection process for this review involved multiple stages. An initial search yielded a large number of publications, which were then screened by titles and abstracts to eliminate irrelevant studies. The remaining studies were subjected to a full-text review, assessing their methodological rigor and relevance to DML and FL.
In the fourth step, we provided a conclusion that summarizes the main findings of the review. This section provides a clear summary of the key takeaways from the review
Finally, we included a reference list of all the studies that were included in the review, following the Springer citation style. This is an important step that allows other researchers to easily find and access the studies that were used in the review.
Overall, the methodology for this review paper on " Introduction to Machine Learning and Decentralized and Collaborative" is a systematic and thorough process that will help to ensure that the review is comprehensive, accurate, and up-to-date

\section{Fundamentals of Machine Learning}
\subsection{Data Preprocessing and Feature Engineering}
Data preprocessing and feature engineering are essential for preparing data for decentralized machine learning and federated learning systems. Decentralized machine learning is being conducted due to the increased privacy threat, so the data tends to remain distributed on different devices or parties. This means being more prepared for any security breach.

\begin{figure}
\centering
\includegraphics[height= 6 cm]{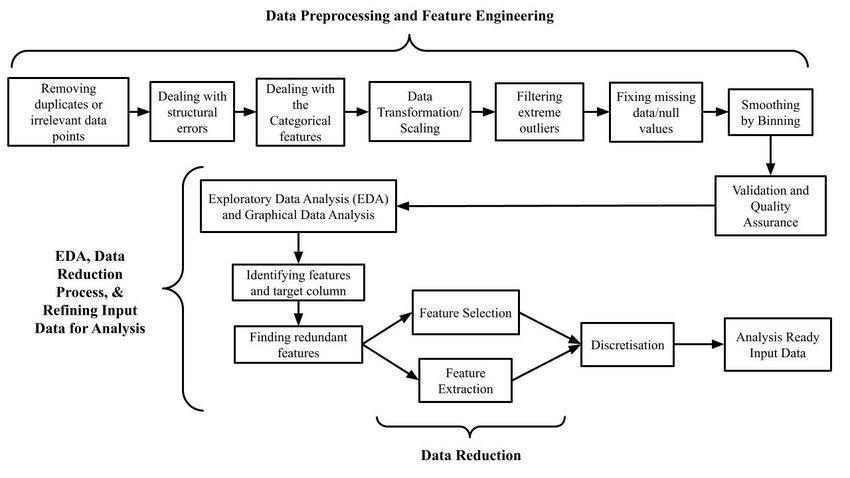}
\caption{Data preprocessing, feature engineering analysis [Source: Md Abrar Jahin, 2023].}
\label{image1}
\end{figure}

\vspace{1em}
The first step of the preprocessing pipeline is to detect and remove missing values or inconsistencies in individual Datasets. Detecting anomalies could also be an outlier, as it could be an outlier if an anomaly is dependent on it. This process is also vital in the final training process since they are outliers, and the data distribution affects the training model. First, they include detecting outliers that might affect the resulting training model, another essential part of preprocessing pipeline features critical to DML.
DML also affects the accuracy of collaborative learning while facilitating the collaborative learning process. Feature selection, furthermore, would be helpful to utilize identity techniques, which are the most useful and would employ the most informative components of a large number of datasets; subsequently, communication overhead would be reduced while the preparation of error rates would be improved. Data transformation could be another technique that could be applied, as it transforms raw features into a suitable model form \cite{ChenLin2014}.
Though traditional techniques such as data preprocessing and feature engineering can be utilized during decentralization, there are some limitations to them. Traditional techniques have data privacy and communication efficiency constraints. Therefore, collaborative filtering approaches where the model only deals with statistical data can sometimes be effective \cite{McMahanetal2017}.

\subsection{Model Selection and Training}
Choosing the appropriate machine learning algorithm and optimizing its training process becomes even more nuanced in decentralized and collaborative settings. Model selection must prioritize communication efficiency and suitability for heterogeneous devices common in DML \cite{Lietal2020}. Lightweight models may often be preferable to computationally intensive approaches. Federated averaging techniques have proven successful in training models across distributed data \cite{McMahanetal2017}.

\begin{figure}
\centering
\includegraphics[width= 4 cm]{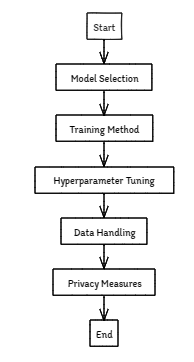}
\caption{Process of model selection and training in decentralized machine learning environments [Source: Self-made]}
\label{image2}
\end{figure}
\vspace{1em}
Hyperparameter tuning is an additional issue. Although much research is dedicated to identifying the optimal setting of the values for the concepts, it can take a lot of work given the multi-device nature of learning in this example. As a result, an open issue remains to be considered: local hyperparameter tuning and rare aggregation or sharing of meta-learning updates may be suitable.
With a frequent lack of identically, independently distributed data across devices, the last issue is non-IID data. Here, the difference in the initial data of devices on which the model was developed also degrades the system's performance. Nevertheless, this issue could be resolved by using data augmentation on individual devices or adjusting personalized models \cite{Konecnyetal2016}.
Moreover, privacy-preserving methodologies like differential privacy will be applied in the training processes. For example, differential privacy helps introduce controlled noise into data, ensuring that individual data points are obscured while maintaining overall statistical properties needed for learning \cite{DworkRoth2014}.

\subsection{Evaluation Metrics and Model Optimization}
In decentralized and collaborative machine learning (ML), evaluating model performance and optimizing model parameters present unique challenges due to the distributed nature of data and computation. 
\begin{table}[ht]
\centering
\caption{Evaluation Metrics and Model Optimization Techniques}
\begin{tabular}{|>{\raggedright}p{6cm}|>{\raggedright\arraybackslash}p{6cm}|}
\hline
\textbf{Evaluation Metrics} & \textbf{Model Optimization} \\ \hline
\textbf{Accuracy and Related Metrics:} Other than accuracy and related metrics, the most fundamental metric that should be covered is when there are imbalanced data sets and varying classes are precision, recall, F1-score, and AUC-ROC with different classes or thresholds help better understand model performance \cite{Powers2020}. & \textbf{Federated Optimization Algorithms:} These are the algorithms used to optimize the model in the case of federated learning. FedAvg, FedProx, and FedNova are the three used in this case to address the data heterogeneity across clients and provide updates to get the model to converge. Moreover, FedNova and FedProx are required to train the model fast while aggregating and solving heterogeneity issues. FedAvg reduces the data utilization among clients by a considerable margin \cite{McMahanetal2017}.\\ \hline

\textbf{Fairness Metrics:} Protecting fairness in collaborative ML among the participants and the distributed systems is critical. Fairness metrics, including demographic parity, equalized odds, individual fairness, and more, are looking to evaluate whether the prediction made by the model is unbiased regarding specific sensitive attributes like race, gender, location, and many views from \cite{Mehrabietal2021}. & \textbf{Hyperparameter Tuning:} Hyperparameter tuning in collaborative machine learning presents challenges due to the distributed nature of data and computation. Active research in this area includes federated hyperparameter optimization (HPO), tuning the hyperparameter across multiple clients, and Bayesian optimization efficiently exploring hyperparameter space \cite{Konecnyetal2016}. \\ \hline

\textbf{Privacy-Preserving Metrics:} Differential privacy, a mathematical framework to quantify privacy loss, is a tool for assessing the privacy guarantees of collaborative machine learning systems. For example, the epsilon-differential privacy metric quantifies the maximum amount of information an adversary may learn about one person’s data from the model’s results \cite{DworkRoth2014}. & \textbf{Personalization:} Given a broad and varied data mix, personalized federated training helps customize a global model to each edge’s client context. Once a global model is trained, it is further updated on the client’s device to optimize it locally for their dataset \cite{Chenetal2020}. \\ \hline

\textbf{Communication Efficiency Metrics:} The communication costs are crucial in decentralized machine learning. It may be bits per parameter update or communication step metrics to evaluate and optimize ML algorithm efficiency in these configurations \cite{Konecnyetal2016}. & \textbf{Robustness and Security:} In joint optimization, the model should optimize for defense against adversarial attacks and data poisoning. For a robust model, adversarial training or robust aggregation will help mitigate the effects of compromised components \cite{Bonawitzetal2017}. \\ \hline
\end{tabular}
\end{table}
The assessment and enhancement of models in decentralized and collaborative ML call for a multi-dimensional strategy considering accuracy, fairness, privacy, communication overhead, and security. By applying suitable metrics and optimization techniques, one can implement robust, fair, and privacy-compliant ML systems capable of harnessing the potential of distributed data and addressing the associated challenges posed by these paradigms.

\subsection{Challenges: Overfitting, Underfitting, and Data Bias}
Even when machine learning is a promising candidate, some parallel issues should be addressed, specifically model testing, a system for generalization, and system bias. The real challenge occurs when a model is trained to fit perfectly the existing features of a dataset, regardless of how signals are separated from noise. An overfitting model performs satisfactorily on the training set but tends to fail on unseen data \cite{Domingos2012}.
On the other hand, underfitting happens when the model is too simple and cannot adequately represent the data, leading to poor performance on all sets (training and testing data) \cite{RaschkaMirjalili2019}. These problems demonstrate the necessity of the trade-off between the complexity of models and the complexity of the task. Moreover, data bias management is an ethical and practical challenge machine learning must confront, considering whether the model's training will replicate existing biases and exacerbate them over time \cite{Mehrabietal2021}.
One primary determinant of biased datasets is historical inequalities, limitations in sampling, or current prejudices in society. Combating data bias necessitates deploying a multi-dimensional approach wherein data collection procedures are carefully examined, fairness-aware algorithms are implemented, and data augmentation techniques are employed \cite{Barocasetal2019}.
With these problems, the dimensions of advanced and combined deployment of machine learning become even more complex. It is essential to consider these identified problems. However, they should not be addressed in isolation because that might affect the integrity of the learning process and the outcome.

\section{Challenges in Traditional Machine Learning}
\subsection{Data Privacy and Security}
The concentration of data in traditional centralized machine learning architectures raises inherent privacy concerns. Sensitive information is ultimately exposed to potential theft, misuse, or breach by unauthorized parties. With the introduction of new privacy regulations such as the GDPR and growing public awareness, the viability of this approach is being seriously questioned. Decentralized and collaborative approaches offer a variety of ways to eliminate and mitigate privacy risks \cite{Abadietal2016}.
In decentralized machine learning cases, data processing is distributed among multiple local devices or participants. Typically, DML reduces the size of centralized datasets. Federated learning is a prime example of possibilities presented by the decentralized approach to privacy preservation. In the FL context, models are trained on decentralized data, and only the resulting model updates or gradients are shared with the central server, ensuring that individuals' raw and personal data are not exposed. However, model inversion attacks highlight persistent risks and the need for further privacy research in FL systems, as in this case, attackers utilize the model updates they receive to attempt to reconstruct personal data \cite{McMahanetal2017}.
Several privacy-enhancing techniques can further strengthen security in decentralized and collaborative learning:
\begin{figure}
\centering
\includegraphics[width= 8 cm]{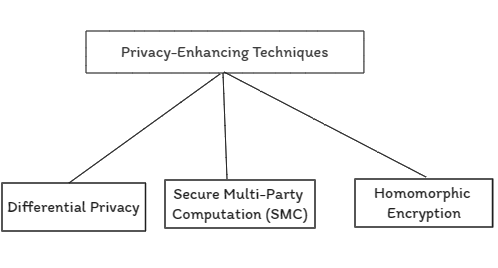}
\caption{The privacy-enhancing techniques in collaborative learning [Source: Self-made]}
\label{image3}
\end{figure}

\vspace{1em}
\noindent\textbf{Differential Privacy:} By strategically adding controlled noise to datasets or model parameters, differential privacy provides guarantees that individual data points cannot be readily identified, even with access to model outputs \cite{DworkRoth2014}.

\vspace{1em}
\noindent\textbf{Secure Multi-Party Computation (SMC):} Secure Multi-Party Computation (SMC) allows multiple parties to perform computations on their combined data without disclosing their private inputs \cite{Lindell2020}. This can facilitate model training or inference on aggregated datasets without compromising individual privacy.

\vspace{1em}
\noindent\textbf{Homomorphic Encryption:} The Homomorphic encryption technique enables computations directly on encrypted data. Models can be trained or used for prediction without requiring decryption, providing a powerful safeguard for data privacy \cite{Gentry2009}.

\subsection{Scalability and Efficiency}
The biggest challenge is addressing the scalability issues of centralized ML models from the classical era in light of the big data phenomenon, where the dataset is exponentially increasing. Large-scale models have computational and storage requirements that exceed current capabilities, meaning that they decrease efficiency. Expanding the infrastructure to the corresponding extent, in order to deliver these growing amounts of energy, becomes a source of high costs and faces physical constraints.
The practicality of decentralized machine learning in addressing the scalability limitations of centralized models is evident. This way, the ML computing tasks and storage can be distributed among a network of devices, allowing for the application of this technology to big data and other computationally intensive applications. In this case, we have an example of parallel processing, which is impossible in a traditional centralized system and leads to a higher level of scalability \cite{ChenLin2014}.
Efficiency also improves through the use of collaborative ML. One of the most prominent examples is federated learning, the FL model, which allows multiple devices on a network to collaboratively train an ML model without relying on a central server \cite{Yangetal2019}. Additionally, channeling data flow through the entire network of devices directly without involving any centralized database enables scaling of this structure, particularly when data transfer is costly or slow, making it more suitable \cite{Lietal2020}.
While evaluating the repercussions of this principle is necessary, decentralization should be given consideration. However, if batch updates of the model are less frequent than exchanging data in raw form, the efficiency of overall communication will still be hindered \cite{Lietal2020}.
Decentralized networks face the challenge of device heterogeneity. The model may assign more significance to devices with more processing power and better connection speed during training. As a result, some devices might not participate fully or at the same level as other devices. Potential solutions include using efficient communication protocols, devising methods to handle device heterogeneity, or even creating hybrids of centralized and decentralized computation, which is also a topic of interest for researchers in the field \cite{Konecnyetal2016}.

\subsection{Bias and Fairness}
While machine learning models are helpful, they also may introduce biases when learning from data. This bias may occur due to unfairness in society, lack of information, or lack of all relevant details \cite{Mehrabietal2021, SureshGutta2021}. If the existing bias is not corrected, it may lead the devices to make unfair decisions, especially regarding critical aspects such as the health care system, financial systems, and the law. Since machine learning is used to draw significant conclusions, ensuring fairness to everyone is essential.
Among the most challenging aspects inherent in machine learning is the development of unbiased systems. Research spans different methods, from modifying the pre-processing steps and rebalancing or reweighting data sample sets to incorporating fairness-aware metrics into the model training process and developing algorithms. Nonetheless, it is vitally important to consider the potential trade-offs between fairness and model accuracy, as this will necessitate careful evaluations of the given circumstances and, therefore, the adoption of context-specific solutions.
Bias and fairness issues are inherent in decentralized and collaborative machine learning (ML) approaches. Additionally, it could be beneficial to include data from more diverse and dispersed sources instead of solely relying on centralized datasets that might have inherent biases. However, when data is non-IID (non-identically and independently distributed) and across different devices in a decentralized system, several challenges arise. Ensuring an equitable representation during collaborative operations, which could call for methods such as data fairness sharing or aggregation approaches, is delicate and may require careful attention. We have presented a comprehensive exploration of bias and fairness issues in decentralized and collaborative learning systems. This ongoing research plays a vital role in our attempts to address these complex challenges.

\section{Decentralized Machine Learning}
A centralized machine learning solution, for all its success, however, is inherently subject to some limitations that seem to increasingly motivate the community towards developing decentralized machine learning systems.

\subsection{Motivations behind decentralized Machine Learning}
 To understand the motivations, authors must assist to understanding the necessity of this type of architectures. Here's a look at some of the key motivations behind decentralizing the machine-learning process:

\begin{itemize}
    \item \textbf{Preserving Data Privacy:} One of the main issues that arise from centralized AI systems is that they require the collection of vast datasets, which means that a lot of information is at risk. Concerning user data, there is a history of breaches, unauthorized use, or misuse regarding this data \cite{Abadietal2016}. Decentralized machine learning (DML) is a framework that helps to keep data localized and limits the data transfer across wide area networks or shared domains where privacy may be a concern.
    \item \textbf{Overcoming Scalability Bottlenecks:}The ML computing and storage requirements can rapidly expand to a point where they are no longer sustainable in the era of exploding data where some big datasets are ten times larger than the entire internet \cite{ChenLin2014}. DML, therefore, does not rely on centralized infrastructure, instead utilizing distributed computing and, as a result, is much more scalable.
    \item \textbf{Addressing Bias and Promoting Fairness:}The datasets developed by the centralized platform are at a higher risk of bias inherent in the population from which the data is collected \cite{Mehrabietal2021, SureshGuttag2021}. DML, alongside data sources that are characterized by broad and diverse definitions, has the ability to minimize bias to a great extent and create a model that is more representative.
    \item \textbf{Enabling Computation in Resource-Constrained Environments:}DML eliminates the device usage barriers for devices with both limited storage and low computational power or network connectivity. This results in the creation of opportunities for machine learning in areas such as the Internet of Things (IoT) and edge computing.
    \item \textbf{Fostering Collaboration with Privacy:}Collaborative ML techniques, such as those seen in federated learning, refer to the fact that, unlike traditional models, they enable organizations to benefit from collective knowledge without having to compromise data privacy. This ushers in a revolution in technologies and business models that help enhance privacy in domains such as healthcare and finance.
\end{itemize}

The migrations to DML are responses to the current problems and future challenges of the data-driven environment. Through decentralized and collaborative approaches, machine learning will be able to extract its strength by overcoming the following obstacles: privacy, scaling, bias, and resource utilization.

\subsection{Differences Between Centralized and Decentralized Machine Learning}
\textbf{Centralized Machine Learning (CML):}
In centralized machine learning, data is consolidated into a single, centralized database. Model training occurs on this aggregated data, and typically, the computers performing these tasks have high computational power and operate on powerful servers or cloud infrastructure.

\vspace{1em}
\noindent\textbf{Decentralized Machine Learning (DML):}
Decentralized ML takes a different approach and distributes computation and data storage across a network of multiple devices or nodes. Often, in DML, data is localized, and models may be trained either individually or collaboratively on each device, without sharing the raw data, but only the updated model parameters.
\vspace{1em}
\begin{table}[ht]
\centering
\caption{Differences Between Centralized and Decentralized Machine Learning}
\begin{tabular}{|>{\raggedright}p{4cm}|>{\raggedright}p{4cm}|>{\raggedright\arraybackslash}p{4cm}|}
\hline
\textbf{Feature} & \textbf{Centralized ML} & \textbf{Decentralized ML} \\ \hline

Data Location & Aggregated in a central repository & Distributed across devices or nodes \\ \hline

Model Training & Single global model on aggregated data & Local models, federated learning, or other collaborative techniques \\ \hline

Infrastructure & Relies on powerful centralized servers & Leverages networks of devices, including edge devices \\ \hline

Privacy and Security & Concentration of data creates privacy risks and a single point of vulnerability & Distributes risk, offers potential for privacy-enhancing techniques \\ \hline

Bias & Prone to reflecting biases in the centralized dataset & Potential to mitigate some biases through access to more diverse data, but requires careful consideration \\ \hline
\end{tabular}
\end{table}

The table above highlights key differences between centralized and decentralized machine learning approaches, particularly in data distribution and model training. In Centralized Machine Learning (CML), data is consolidated into a single location, eliminating data distribution challenges. However, this centralization can lead to privacy risks and scalability bottlenecks, issues that Decentralized Machine Learning (DML) addresses by distributing data and model training across multiple devices.
On the other hand, DML allows data to be localized and uses distributed computation. However, it introduces additional difficulties as data is often distributed and cannot be assumed homoscedastic; thus, it needs to be adjusted to ensure that it is suitable for training and that the model can converge. Finally, the DML introduces new risks, such as model poisoning, that should be mitigated due to multiple heterogeneous devices scattered across numerous households.

\subsection{The Shift to Decentralization}

\textbf{Centralized ML:}
First, traditional centralized machine learning frameworks rely on massive datasets funneling into a central pool to function. This raises privacy issues, as any data breach or unauthorized access could compromise users’ private data. Second, CML faces difficulties scaling to utilize large datasets due to bottlenecks. Since CML struggles to work with exceedingly large datasets, it has difficulty training complex models.

\vspace{1em}
\noindent\textbf{Emergence of DML:}
DML, as a decentralized approach to resolving the limitations of traditional ML, is here to stay.
\begin{itemize}
    \item \textbf{Federated Learning:} This empowers models to run in a distributed environment and train on data distributed across various devices or institutions. Only model updates, not raw data, are shared, thus ensuring privacy preservation.
    \item \textbf{Split Learning:} It divides neural network models between local and centralized servers. Consequently, the training can be performed on sensitive data locally, increasing privacy and efficiency.
\end{itemize}

\subsubsection{Applications of DML} There are new exciting opportunities and unique advantages in decentralized machine learning:
\begin{itemize}
    \item \textbf{Privacy-Sensitive Healthcare:} With the help of DML, doctors or clinics may collaborate on medical research based on their patient’s medical histories or MRI images without sharing the raw data. This can lead to the rapid growth of diagnostic tools and personalized treatments in the health field.
    \item \textbf{IoT Sensor Networks:} DML allows devices to train on-device and make decisions in real-time in IoT applications, where sensors must react quickly without exhausting energy or communication bandwidth.
\end{itemize}

\begin{figure}
\centering
\includegraphics[height= 6 cm]{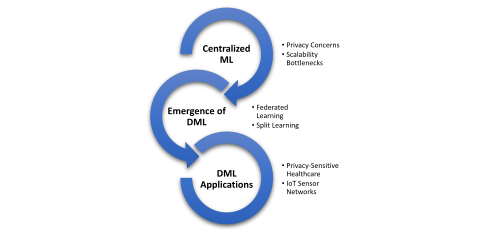}
\caption{The Shift to Decentralization [Source: Self-made]}
\label{image4}
\end{figure}

\subsection{Advantages of Decentralization}
The shift towards decentralized machine learning architectures offers several compelling advantages over traditional centralized approaches. These advantages address core limitations and drive future innovation in the field:

\begin{table}[ht]
\centering
\caption{Advantages of Decentralized Machine Learning}
\begin{tabular}{|>{\raggedright}p{4cm}|>{\raggedright\arraybackslash}p{8cm}|}
\hline
\textbf{Advantages} & \textbf{Description} \\ \hline

Enhanced Privacy and Security & Decentralization, most importantly, enable the elimination of such a process by removing the centralized location or warehousing of sensitive and confidential information. Models trained on local data also reduce the surface area of attack since personal privacy may only be inferred from the device with access to its information. This is the case with environments where the underlying data is highly personal, such as healthcare or finance. \\ \hline

Improved Scalability and Efficiency & One crucial benefit of decentralized machine learning is the possibility of training a model over thousands of geographically distributed devices and edge networks. This approach essentially allows tapping into vast previously underused computing capacity while training on a significantly more extensive and diversified data set, which might eventually lead to superior generalization and performance \cite{Niknametal2019}. \\ \hline

Increased Resilience and Fault Tolerance & Eliminating single points of failure make the entire system more resilient to accidental node malfunction or targeted attacks. If one of the devices becomes inoperative, the remaining network could continue learning, making the system more available and robust \cite{Huangetal2021}. \\ \hline

Reduced Bias and Increased Fairness & Centralized approaches to training lead to data-set bias, which raises the possibility of the distribution of model training across multiple diverse datasets held at the source. Consequently, decentralization produces fair machine learning \cite{Lietal2020}. \\ \hline

Enhanced User Control and Autonomy & Additionally, decentralized frameworks introduce people with greater control over their data and the opportunity to establish whether and how it leads to a training model. Thus, this aspect supports the human rights orientation to the framework \cite{Hardetal2018}. \\ \hline
\end{tabular}
\end{table}

\section{Collaborative Machine Learning Approaches}

Machine Learning approaches provide many stakeholders to work collaboratively in terms of training models 

Collaborative Machine Learning approaches enable multiple parties to work together on training models without sharing raw data, enhancing both privacy and utility. By distributing model training across multiple sources, these methods mitigate privacy risks while utilizing diverse data. Federated learning is a key approach where each party trains a model locally and shares only model updates, preserving data confidentiality. Such approaches are valuable in sensitive fields like healthcare and finance, where data privacy is paramount. With frameworks like Zero Trust, collaborative Machine Learning ensures robust security and efficient communication between participating entities.

\subsection{Core Principles in the }
CollaConcerned Domainborative machine learning is a class of techniques that enables training machine learning models across distributed datasets without aggregating the raw data in a central location. CML approaches, such as federated learning, were used as a core technical mechanism in several ICLR papers in the field.
Collaborative machine learning offers the following advantages: data locality and privacy preservation. Data can remain on the originating device or local network. Instead of sharing raw data, only model updates or gradients are shared, significantly minimizing privacy and security risks associated with data transfer \cite{Bonawitzetal2017}.
In collaborative model training, multiple parties participate in a joint training process without sharing their datasets. This collaboration can be led by a server or adopt a decentralized communication structure; in both scenarios, it aims to develop a global model that leverages the strengths of all individual datasets \cite{Yangetal2019}.
Heterogeneity tolerance is essential in CML systems, as they must work in diverse environments where participants have various computational capabilities, data distributions, and varying network connectivity. Robust CML algorithms can handle these heterogeneities to make collaboration inclusive and effective \cite{Kairouzetal2019}.
Furthermore, CML must prioritize security and integrity. Privacy guarantees alone are not sufficient; CML must also address potential attacks from malicious participants trying to corrupt the model or infer sensitive information from its outputs. Techniques such as differential privacy, secure aggregation, and homomorphic encryption can be combined to enhance security \cite{Gentry2009}.

\subsection{Federated Learning Process and Architecture}
\textbf{Federated learning process:} Federated learning is a popular method of cooperative machine learning by various participants. FL allows multiple players to grow a model collectively without sharing private data by learning a process coordinated by a central server manifested in the whole process.

\begin{figure}
\centering
\includegraphics[width= 12 cm]{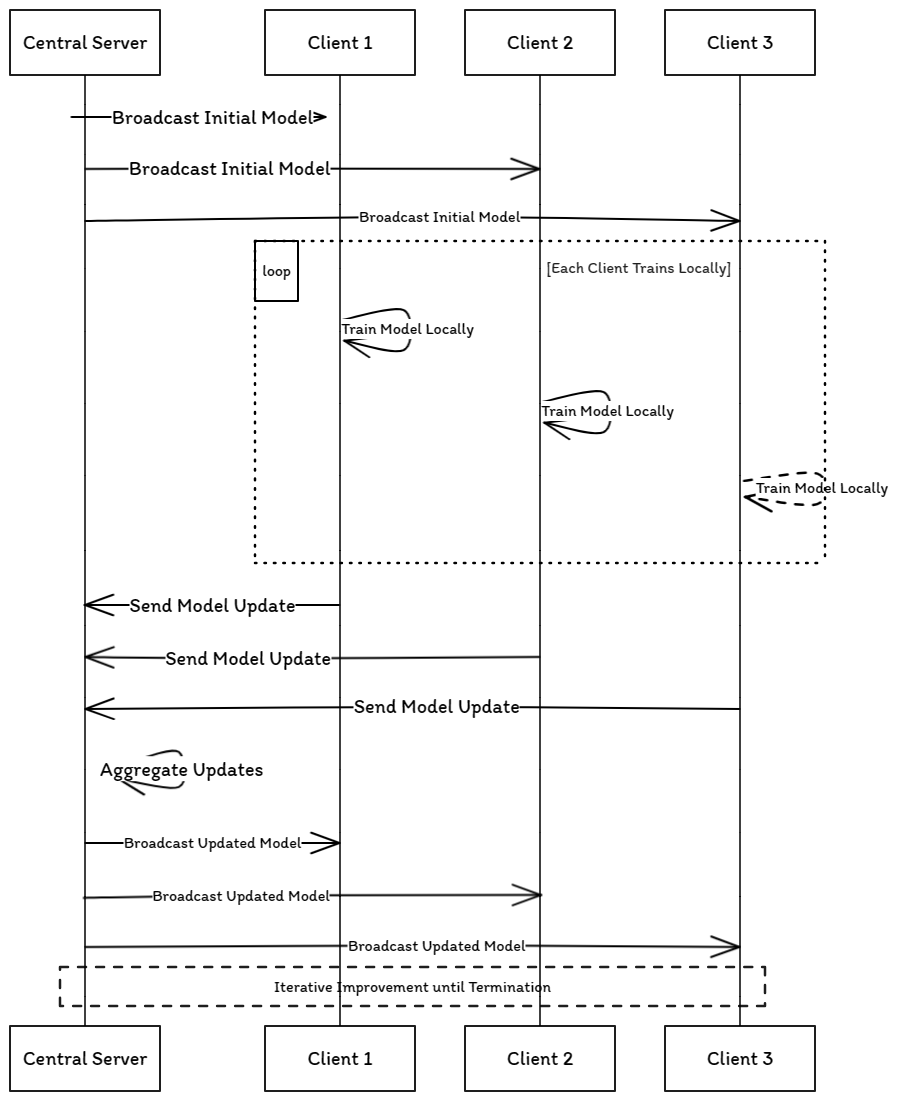}
\caption{Federated learning process [Source: Self-made]}
\label{image5}
\end{figure}
\vspace{1em}
\noindent\textbf{Initial model distribution:} The central server conveys the initial global model to each participating client, device, or local server. The initial model could include a weak model or a model that was trained using a public dataset \cite{McMahanetal2017}.

\vspace{1em}
\noindent\textbf{Local model training:} Each client feeds the local version of the model over its local data. In this case, training refers to processing the model through the local data and locally adjusting the model’s weights and biases \cite{Yangetal2019}.

\vspace{1em}
\noindent\textbf{Local model updates:} Clients send the model’s updates to the central server. A client typically retrieves the model, calculates the differences between the initial model and the model trained on the client’s data, and then sends these updates \cite{Bonawitzetal2017}.

\vspace{1em}
\noindent\textbf{Global model aggregation:} The central server aggregates the numerous model updates sent from the clients and produces a new global model \cite{Kairouzetal2019}.

\vspace{1em}

\noindent\textbf{Global model iteration: } The updated global model is distributed to the clients, and the entire process is repeated until the model achieves the desired performance \cite{Lietal2020}.

\begin{figure}
\centering
\includegraphics[width= 8 cm]{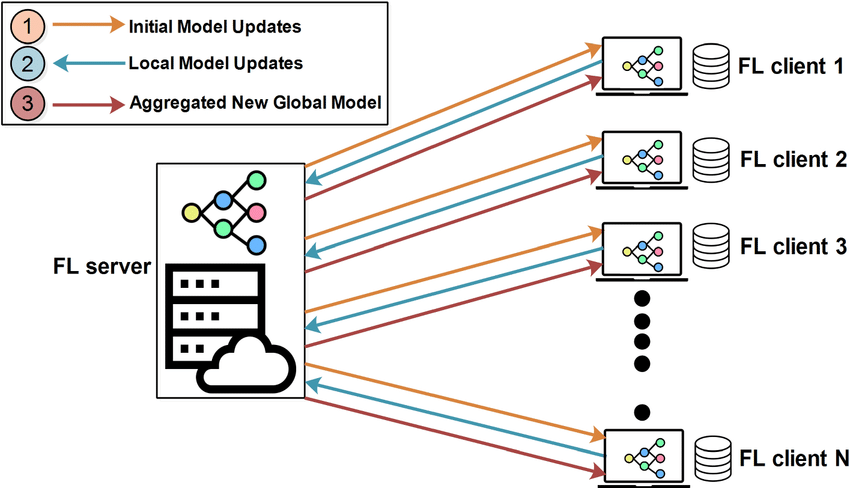}
\caption{General working process of federated learning}
\label{image6}
\end{figure}

\subsection{ Applications of Collaborative ML}
Collaborative machine learning, specifically through federated learning and its variations, finds applications across a diverse range of industries and domains. Some key areas include:

\begin{figure}
\centering
\includegraphics[width= 12 cm]{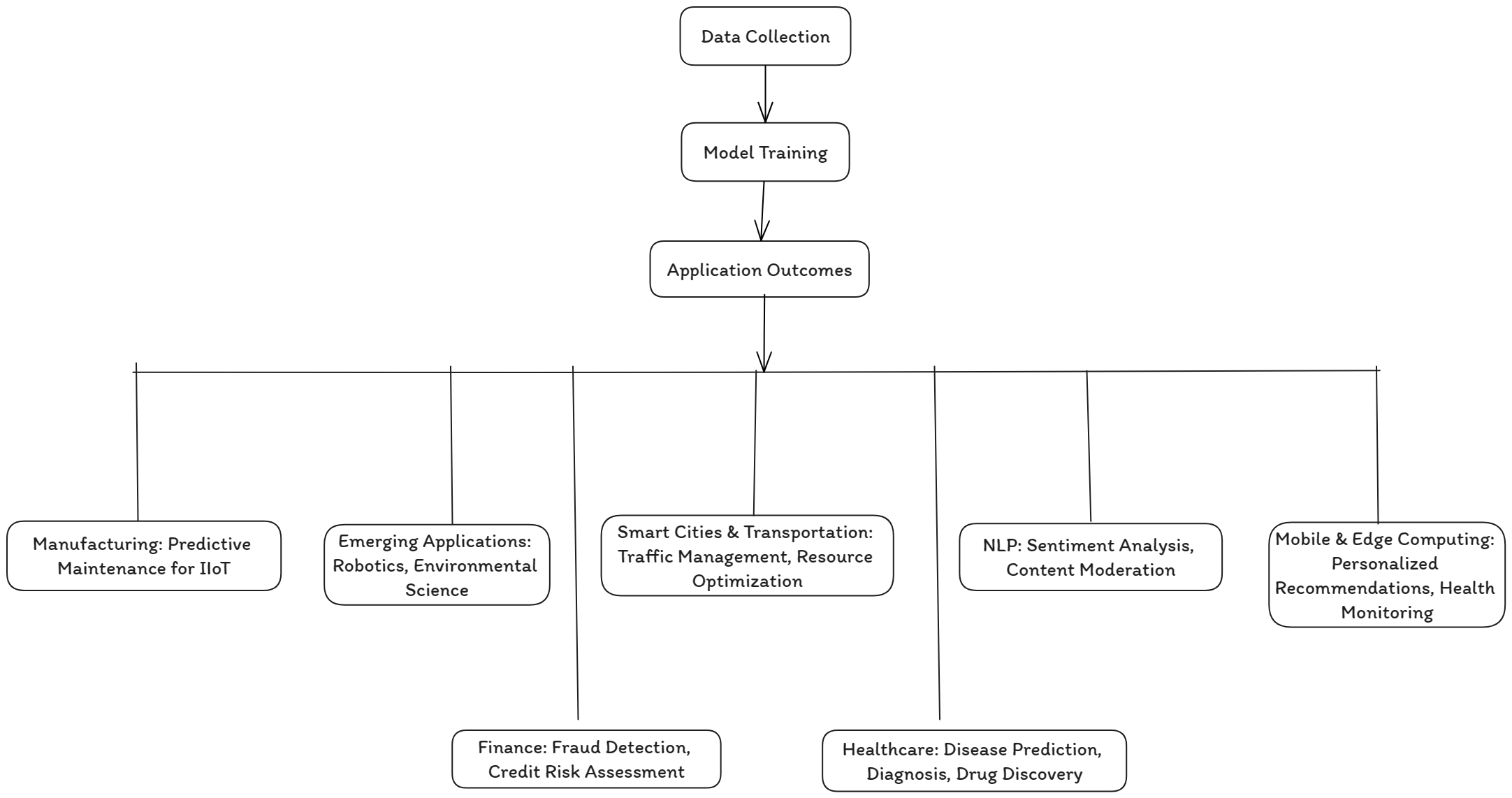}
\caption{Applications of Collaborative ML [Source: Self-made]}
\label{image7}
\end{figure}

\begin{itemize}
    \item \textbf{Healthcare:} Through collaborative model training between numerous healthcare institutions and a wide variety of data sources, disease predictions and diagnoses will be more accurate, while patients' data privacy remains guaranteed \cite{Riekeetal2020}. Other examples of biotechnology where the decentralized approach is applicable include genome sequencing, drug discovery, and precision medicine \cite{Savazzietal2020}.
    \item \textbf{Finance:} By working together, financial institutions can innovate in fraud detection and credit risk assessment models without disclosing sensitive customer data \cite{Niknametal2020}. Additionally, it is easier to detect money laundering patterns since machine learning models can be trained on datasets obtained from multiple entities on a large scale \cite{Bagdasaryanetal2020}.
    \item \textbf{Manufacturing and Industry 4.0:} The technique of federated learning allows for predictive maintenance of industrial Internet of things (IIoT) devices by aggregating knowledge from equipment across different environments. This contributes to more efficient fault detection and the prevention of unexpected downtimes \cite{Savazzietal2020}.
    \item \textbf{Smart Cities and Transportation:} Collaborative ML has the potential to optimize traffic flows using distributed sensors by collaborating on predictions, resource allocation, and faster incident response. Data integration between different transportation entities does not raise data ownership issues \cite{Lietal2021}.
    \item \textbf{Mobile and Edge Computing:} Collaborative processes are instrumental in enabling on-device learning, in which smartphone or wearable user data improves the generalization of predictive models for user health monitoring and personalized recommendations. This method maintains data privacy and allows for efficient model updates.
    \item \textbf{Natural Language Processing (NLP):} Collaborating language models can be trained using distributed text data from users or organizations, thereby enhancing the performance of sentiment analysis, content moderation, and language translation \cite{Chenetal2020}.
    \item \textbf{Robotics and Autonomous Systems:} Collaborative methods could be used in training robot swarms or multi-robot systems in diverse environments, sharing knowledge across the system while optimization for specific adaptation is handled locally.
    \item \textbf{Environmental Science:} Data from sensors and research institutions located across various geographical locations can be used collaboratively, enabling climate modeling and natural disaster prediction through global knowledge sharing without exposing raw data to numerous parties.
\end{itemize}

\subsection{Challenges and Solutions: Communication Efficiency, Heterogeneity, and Security}
Collaborative machine learning (CML) and decentralized machine learning (DML) hold immense promise for privacy-preserving, scalable, and fair model training. However, their practical implementation is fraught with challenges that necessitate innovative solutions.

\begin{itemize}
    \item \textbf{Communication Efficiency:} Communication overhead is a severe problem in collaborative settings as model updates are passed between many devices or people. This worsens with the increasing number of models and updates and the low bandwidth or unreliable connections in some operating environments. To mitigate this issue, researchers are investigating several approaches. These approaches include model compression, whereby one only sends the essentials of the model and gradients, such as gradient reconstruction. Gradient quantization seeks to shrink the bandwidth used to send gradients by reducing their precision. One can also use asynchronous communication protocols and local model averaging to reduce the number of communications with the server \cite{Lietal2020}. Other solutions include personalized federated learning.
    
    \item \textbf{Heterogeneity:} As previously mentioned, DML devices are inherently different, and their distributions complicate the training process. Uneven implementation of devices in training can lead to biased models and slow convergence. The heterogeneity of DML devices is solved by developing FL algorithms that adapt to such conditions. According to Li et al., federated learning methods like FedProx and FedNova can adjust client learning rates or model updates to their local conditions \cite{Lietal2020}. Moreover, personalized FL makes it possible to change local models while still contributing to improving the global model.

    \item \textbf{Security:} Decentralized systems are more prone to security issues than centralized ones. CML may result in model poisoning, where corrupted data or model updates are injected, and adversarial attacks that exploit vulnerabilities. Security issues can be fixed using differential privacy approaches to control the addition of noise to model updates, meaning it would be hard to understand sensitive information from individual submissions \cite{DworkRoth2014}. Secure multi-party computation and homomorphic encryption also protect part data during collaborative computations \cite{Lindell2020}. Zero Trust’s security model has been gaining popularity in DML defense lately \cite{Stafford2020}.

\end{itemize}

Overcoming these challenges is crucial to unlocking the promise of cooperative and distributed machine learning. The field is already advancing communication-efficient algorithms, heterogeneity-aware optimization strategies, and resilient security measures, giving rise to a mature and reliable CML and DML-enabled future that is also secure, scalable, and strong. They can be hailed as intelligent new-generation applications deployed in various markets, transforming how we use the data on which they are founded to discover and make decisions while maintaining intellectual privacy and equity.

\section{Zero Trust in Machine Learning}
\subsection{Concept of Zero Trust Security Model}
The Zero Trust security model is a distinct departure from traditional security models based on a perimeter. Whereas older models operated under the assumption that elements inside a given network perimeter could be trusted, Zero Trust assumes that no entity within is inherently secure. This is especially relevant when considering machine learning, as data sources and user interactions may span various organizations.

\vspace{1em}
\noindent\textbf{Central to the Zero Trust model is the following principles:}

\begin{figure}
\centering
\includegraphics[width= 12 cm]{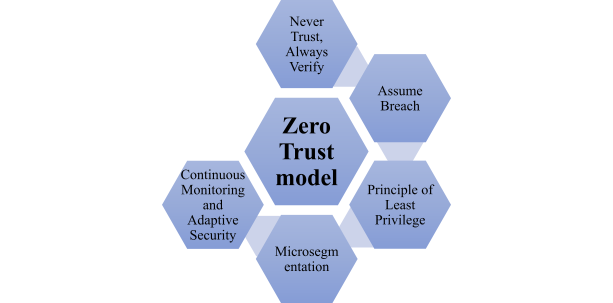}
\caption{Zero Trust Security Model [Source: Self-made]}
\label{image8}
\end{figure}

\begin{enumerate}
    \item \textbf{Never Trust, Always Verify:} Although the Zero Trust principles permit any access request (be it a user, device, or app), only such a request will be considered valid after authentication, authorization, and continuous verification. This applies whether the request is made from inside or outside the organization's network perimeter.
    
    \item \textbf{Assume Breach:} The model, in turn, advocates for assuming that a network or system compromise is highly likely or has already occurred. Hence, it leaves no room for complacency and enforces rigorous, fine-grained access controls.

    \item \textbf{Principle of Least Privilege:} Access rights granted to users and devices are restricted to the absolute minimum levels of access permissions needed to fulfill their intended functions. This principle minimizes the impact of compromised credentials and consequently reduces the attack surface.

    \item \textbf{Microsegmentation:}  The application of Zero Trust encourages the partitioning of networks into smaller, isolated segments, each with specific access policies and security checks tailored to that segment. This prevents attacks from spreading and effectively restricts the lateral movement of malicious actors within the network.

    \item \textbf{Continuous Monitoring and Adaptive Security:} Real-time monitoring of user activity and device behavior is a critical component of Zero Trust architectures. Utilizing data analytics and behavioral analysis services provides opportunities to detect and flag anomalies, enabling adaptive and effective responses.

\end{enumerate}

Growing networks with dispersed IT environments, the widespread adoption of cloud services, remote workforces, and the diffusion of endpoint (particularly IoT) devices threaten the effectiveness of the traditional "castle-and-moat" security approach. Applying such tactics as authentication and gaining users' confidence centered on assuming the compromised nature of the network would provide a more effective and scalable security solution for a modern complex infrastructure.

\subsection{Application of Zero Trust Principles in Machine Learning} The Zero Trust security model, rooted in the principle of "never trust, always verify," is increasingly being recognized as a critical framework for securing machine learning systems, particularly in decentralized and collaborative settings. Applying Zero Trust principles to ML involves rethinking traditional security assumptions and adopting a proactive, layered approach to protect data, models, and the overall learning process.

\begin{figure}
\centering
\includegraphics[width= 12 cm]{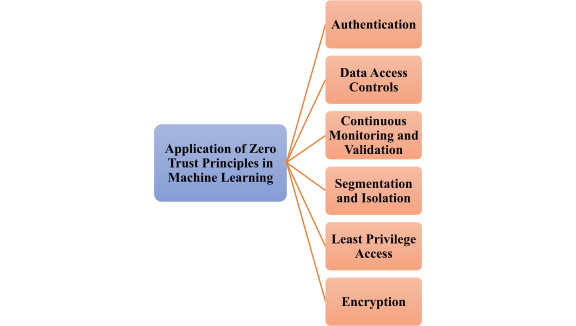}
\caption{Application of Zero Trust Principles in Machine Learning [Source: Self-made]}
\label{image9}
\end{figure}

\begin{enumerate}
    \item \textbf{Authentication:} All devices, users, or entities navigating the learning process should thus be authenticated and granted access to Zero Trust ML. This can be achieved by verifying and authorizing these participants based on the least privilege principle, where they can only do what is necessary for their role. Solid authentication mechanisms using multi-factor authentication and cryptographic protocols can reduce the exposure to unauthorized parties \cite{Stafford2020}.
    
    \item \textbf{Data Access Controls:} Zero Trust in the ML framework also implies more granular data access controls. Since the dataset, when training federated, may be dispersed in the devices of respective entities, access controls should thus be imposed from the source destination. Only these entities should be authorized to access and operate data through encryption, ACLs, and data anonymization techniques such as differential privacy \cite{DworkRoth2014}.

    \item \textbf{Continuous Monitoring and Validation:}  In Zero Trust ML, the behavior of all the players within the learning process should be monitored and validated at all times. This includes tracking data inputs, model updates, and even communication to determine whether any suspicious activity is evident. To mitigate such potential threats, learning entities can develop machine learning models to model offenders and their behavior, such as poisoning or data breaching. Anomaly detection and intrusion detection are possible tools for mitigating under real-time scenarios.

    \item \textbf{Segmentation and Isolation:}  Zero Trust secures the segmentation and isolation nature. Federation learning should thus use a different communication channel for various entities or isolate the training model from surrounding elements. The limited potential will require a smaller sandbox to safeguard against potential attacks.

    \item \textbf{Least Privilege Access:} In this context, all players require limited access tools to facilitate their access roles and minimize harm if one data entry is affected. This possible tool may be role-based access control (RBAC) or attribute-based access control (ABAC) for the least privilege.

    \item \textbf{Encryption:} The roles in Zero Trust ML may defer a lower attack role by the exchange protocol motioning to servers or from a server to the exchange itself. The possible role is to secure the saved model or update-level encryption to reduce the unauthorized update or crime action. The option-given opportunity may suggest that specific data proves transition is possible \cite{Gentry2009}.

\end{enumerate}

By incorporating Zero Trust principles into the design and implementation of machine learning systems, organizations can enhance the security and resilience of these systems, especially in decentralized and collaborative settings. This approach is essential for protecting sensitive data, ensuring the integrity of models, and building trust among participants, ultimately enabling the broader adoption and deployment of ML technologies in various domains.

\subsection{ Enhancing Security and Trust in Federated Learning Systems}
Federated learning, while solving the problem of user data privacy, creates specific security challenges. Several techniques can be used to generate trust in collaboration models. These techniques are protecting model updates that are exchanged during training. One such approach that McMahan et al. can use is differential privacy. This technique adds noise to the model updates; therefore, the individual contributes to training. As a result, the clandestine information on gradients of selected observations becomes challenging, as they need to reflect the actual model status. For an even higher degree of privacy, homomorphic encryption by Riazi et al. (2020) can be used. This technique allows computations to be performed on encrypted data, effectively guaranteeing the safety of individual contributions. However, homomorphic encryption is computationally expensive. Secure multi-party computations by Bonawitz et al. can be used instead, as they offer the functionality for securely aggregating model updates without revealing individual contributions \cite{Bonawitzetal2017}

\vspace{1em}
The fundamental problem with FL, after mitigating the risks of decentralized model training, is the possibility of poisoning attacks. This occurs if a user purposefully updates their local model with false data, thus sabotaging the final iteration of the global model. Anomaly detection and response systems, as well as robust aggregation algorithms, can help to deal with this issue \cite{Blanchardetal2017}. Finally, even when the model is fully trained, its use presents the danger of inference threats. The model can be used to extract sensitive information. This is especially harmful for non-federated models, as their release ensures more intelligence to the party using it. To prevent this, models with an added degree of differential privacy can be trained, or knowledge distillation can be added into the training process, with a resulting “student” model being offered to the public. For accountability and security, it is also essential to have clear auditing trails of model updates and communication logged in total. Overall, Zero Trust is the philosophy that best complements the model by continuously verifying client identities and following the principle of least privilege by unquestioningly offering model requests without logical implementation after auditing the source of the update.

\subsection{Future Directions and Research Opportunities
}
Decentralized machine learning and more collaborative approaches may transform machine learning. Research is in progress to remedy the constraints and opportunities inherent in current systems.
One area of study would be enhancing federated learning technology's privacy and security assurances. Research into more advanced homomorphic encryption schemes may decrease the computational cost of federated learning \cite{Gentry2009}. Knowledge can be shared while data confidentiality is guaranteed.
Moreover, more privacy-friendly approaches to FL dynamics should be studied to infer valuable models from the large datasets that are consistently generated while preserving individuals from data spills \cite{DworkRoth2014}. Even when implementing decentralized systems, researchers must devise algorithms that work effectively across the different kinds of data in federated learning settings. They would need to learn at a high level of performance. Personalization and knowledge transfer targeted to federated learning are potential options \cite{Kairouzetal2019}.
Fairness and bias reduction will be essential to building decentralized marketplaces where several parties share data and models. This must be done to avoid accumulating bias. Decentralized research could investigate DSML use cases and the integration of blockchain technologies. It would give insight into secure model updates and audit logs \cite{Niknametal2020}. We may need to determine how the platform encourages gathering data without sacrificing model quality.
With the rise of IoT and edge gadgets, federated learning that is resource-efficient and sustainable learning is also in high demand \cite{Chenetal2020}. Finally, research such as ethical DSML, the DLM’s companion, and market compliance with consumer privacy are essential. Both are intended to improve data rights and governance awareness. The future demands addressing actual difficulties robustly, responsibly, and ethically \cite{Barocasetal2019}.

\section{Conclusion and Outlook}
This chapter involved a comprehensive history of developing machine learning (ML) systems restricted by the conventional centralized model and alternatives in decentralized and collaborative approaches. We have outlined the core concepts of machine learning and specified the importance of data pre-processing and the type of model applied in the training and testing phase. In addition to describing positive machine learning capabilities, we have discussed the problems, such as data over-fitting, under-fitting, and bias, which constantly restrain the machine learning capability. There have been noted both remarkable efficiencies of centralization up until now and several crucial drawbacks. Privacy problems, limited scalability, and the potential to reinforce biases have motivated the new tendency to decentralize. Federated Learning is another form of decentralized machine learning that has presented the most progress and drawn the most attention of researchers. This form permits the unleashing of the power of decentralized data while keeping personal data private since the FL form allows the development process to be collaborative.

\vspace{1em}
In contrast, both the model and data are spread to third parties. Federated learning has launched progress and revealed new opportunities in healthcare and finance. However, new challenges arise when the system is decentralized, such as communication inefficiencies and heterogeneity in devices and data. Fortunately, the researchers have already given strategies to address this challenge, including model compression and efficient asynchronous communication protocols. Federated optimization and personalized federated learning are also going through development to avoid an inadequate heterogeneous environment. There are additional methods, including differential privacy, secure multi-party computation, and homomorphic encryption methods, to build security and trust in collaboration and agreement among systems. Security involves the Zero Trust model – never trust and always verify, and in partnership with machine learning, several applicable policies can be applied. The first policy is to validate before authorizing entry and continued monitoring. The second is to share classification. The third is to restrict entry to a specific range. The fourth is to access minimal power and, eventually, provide service with strong encryption. These strategies guarantee that private data are protected and only suitable knowledge is used.

\vspace{1em}
With the ongoing development and growth of machine learning, it is impossible to underestimate the role of a distributed and collaborative approach in its evolution. The evolution of distributed machine learning, currently known as DML, and federated learning, in particular, have proven their potential to reshape the efficiency and performance of ML-related data use with a central focus on privacy and developing bias-free environments.

\vspace{1em}
\noindent\textbf{Top trends and predictions for machine learning’s future include:}

\begin{itemize}
    \item \textbf{Increased adoption of federated learning (FL):} The future of machine learning will witness the rise of FL. It has the potential of privacy-first distributed data science to reveal secrets hidden in faraway servers. The technology is expected to take over the global market shortly. This results from the consumer demand that has continued to develop FL algorithms and optimization techniques and enhance communication protocols to make them more scalable and efficient \cite{Niknametal2020}.

    \item \textbf{The rise of edge computing:} The future of DML will support the increase in edge computing as more and more edge devices lead to DML enabling training and inference on a single device with reduced data traffic. Continuous large sets of data also mean that faster and customized machines will be developed for specific purposes, especially in healthcare, IoT, and mobile sectors \cite{Chenetal2020}.

    \item \textbf{Focusing on explainability and equity:} There is a trend towards a demand for ML working models with transparent and user-understandable design. This is because more focus is expected to be put into boosting the capabilities of explainable AI to expose the inner workings of an algorithm and to hold the maintainers accountable and open to questioning \cite{Barocasetal2019}.
    
    \item \textbf{Utilizing blockchain technology:} Blockchain can boost confidence between collaboration parties due to its large scale of distribution and trust ledger settings. Blockchain will likely be integrated into DML ecosystems to give security updates, audit trails, and incentive systems, among other vital applications \cite{Niknametal2020}.

    \item \textbf{Establishing and practicing ethical and regulatory guidelines:} Machine learning has developed quickly, and although change is good, it is essential that conducting methods are ethical and fair. Ways to remove bias, data control, and transparency will be key talking points and discussions in policy-making. The trends are set to make machine learning models more rigorous, transparent, and ethical, leading the way for privacy and fair technology in the future \cite{Mehrabietal2021}.

\end{itemize}


\bibliography{sn-bibliography}

\end{document}